\documentclass{ecai}
\usepackage{times}
\usepackage{graphicx}
\usepackage{latexsym}
\usepackage{multirow}
\usepackage[utf8]{inputenc}
\usepackage{hyperref}
\usepackage{url}
\usepackage{booktabs}
\usepackage{amsfonts}
\usepackage{microtype}
\usepackage{times}
\usepackage{soul}
\usepackage{url}
\usepackage[small]{caption}
\usepackage{graphicx}
\usepackage{amsmath}
\usepackage{booktabs}
\usepackage[usenames, dvipsnames]{color}
\usepackage{tabularx}
\usepackage{graphicx,subcaption}
\usepackage{varwidth}
\usepackage{cancel}
\usepackage[normalem]{ulem}
\usepackage{scrextend}
\usepackage[linesnumbered,ruled,vlined]{algorithm2e}
\usepackage[shortlabels]{enumitem}

\SetCommentSty{mycommfont}

\SetKwInput{KwInput}{Input}                
\SetKwInput{KwOutput}{Output}              
\newcommand{\vect}[1]{#1}
\newtheorem{definition}{Definition}

\DeclareMathOperator*{\argmax}{arg\,max}

\begin{document}

\title{Structure Matters: Towards Generating Transferable Adversarial Images
}

 \author{Dan Peng\institute{Harbin Institute of Technology (Shenzhen),
China, email: \{pengdan,luolinhao\}@stu.hit.edu.cn, zhangxiaofeng@hit.edu.cn. Xiaofeng Zhang is the corresponding author.}
\and 
Zizhan Zheng  
\institute{Tulane University, USA, email: zzheng3@tulane.edu.}
\and 
Linhao Luo\footnotemark[1] 
\and
Xiaofeng Zhang\footnotemark[1]
 }
 
\maketitle

\bibliographystyle{ecai}


\begin{abstract}
Recent works on adversarial examples for image classification focus on directly
modifying pixels with minor perturbations.
The small perturbation requirement is imposed to ensure the generated adversarial examples being natural and realistic to humans, which, however, puts a curb on the attack space thus limiting the attack ability and transferability especially for systems protected by a defense mechanism.
In this paper, 
we propose the novel concepts of structure patterns and structure-aware perturbations that relax the small perturbation constraint while still keeping images natural. The key idea of our approach is to allow perceptible deviation in adversarial examples while keeping structure patterns that are central to a human classifier. Built upon these concepts, we propose a \emph{structure-preserving attack (SPA)} for generating natural adversarial examples with 
extremely high transferability.
Empirical results on the MNIST and the CIFAR10 datasets show that SPA exhibits strong attack ability in both the white-box and black-box setting even defenses are applied. Moreover, with the integration of PGD or CW attack, its attack ability escalates sharply under the white-box setting, without losing the outstanding transferability inherited from SPA. 

\end{abstract}

\section{INTRODUCTION}\label{sec:introduction}
Deep neural networks (DNNs) have achieved
phenomenal success in computer vision by showing superior accuracy over traditional machine learning algorithms. However, recent works have demonstrated that DNNs are vulnerable to adversarial examples that are generated for malicious purposes~\cite{szegedy2013intriguing,2014arXiv1412.6572G}.
This observation has raised serious concerns on the robustness of 
the state-of-the-art DNNs and limited their applications in various security-sensitive applications~\cite{eykholt2017robust,malware_attack}. 

Generally speaking, adversarial examples can be any valid inputs to machine learning models that are intentionally designed to cause mistakes~\cite{elsayed2018adversarial}.  
Although intuitively, an input is valid as long as it is natural and meaningful to human eyes, how to quantify this formally in the objective function is challenging. 
 For object recognition, we rely on human labelers to obtain the ground truth labels, which are unknown to the attacker before natural adversarial examples are generated. 
To bypass this dilemma, a common attack strategy is to start with a clean image 
where the ground truth label is already known and modify it so that the new image is natural and semantically
similar to the original image while its output label differs from the ground truth label of the clean image.

 \begin{figure}[!t]
 \centering
 \includegraphics[height=0.8 in,width=0.45\textwidth]{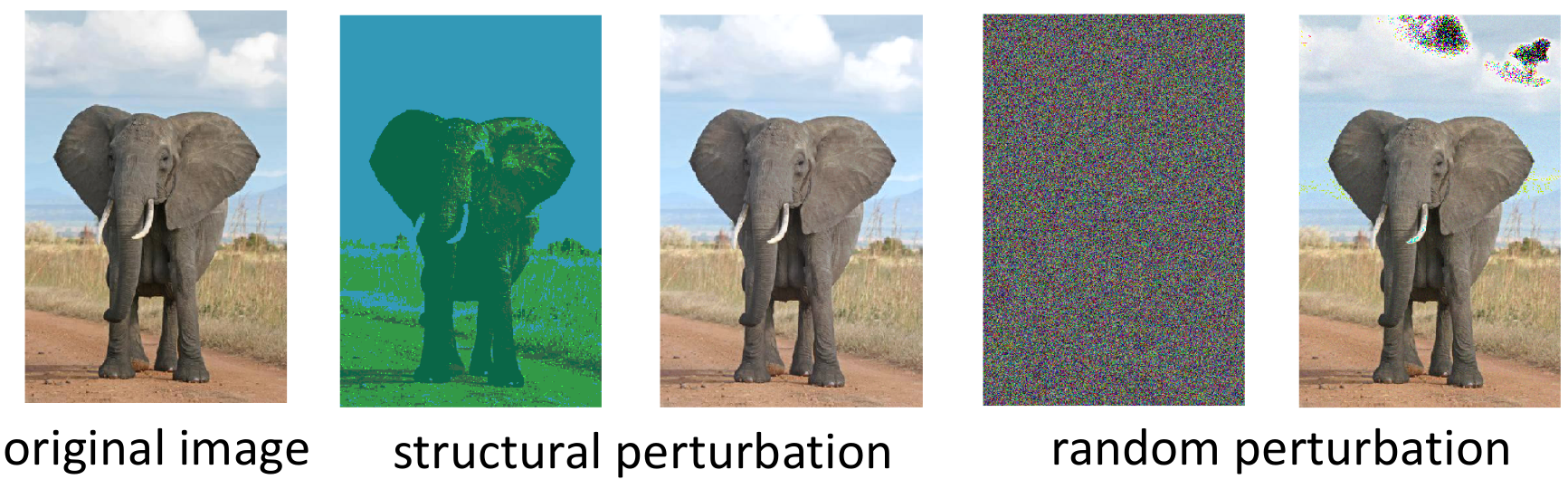}

 \caption{An example of structural perturbations. The second and fourth images are the structural and random perturbations, respectively, under the same maximum allowed distortion size. The two images are scaled (with pixel values enlarged by a factor of 10) for clear illustration. The third and fifth images are the adversarial images generated from the original image with the corresponding perturbations added. Structural adversarial examples are more natural than adversarial examples generated by adding 
 random perturbations.
 }
 \label{fig:intro}
 \end{figure}

A simple approach for obtaining natural adversarial examples and ensuring semantic
 similarity that has been intensively studied in the literature is to introduce small perturbations into pixels such that the distortion between the adversarial example and the original image
is humanly imperceptible~\cite{2014arXiv1412.6572G,su2018attacking,chen2017ead,carlini2017towards}.  
Note that the perturbation considered in these works is typically unstructured as random noise, thus only very small perturbations can be allowed to be superimposed onto images; 
Otherwise, the large unstructured distortion will destroy the semantics of the original image and further make the generated image unnatural and less meaningful. We highlight that the small perturbation requirement is not a necessity. The ultimate goal 
is to ensure the generated images being both natural and realistic to human eyes. 
However, the small perturbation size leads to the low transferability of these perturbation-based attacks (see the detailed analysis in Section~\ref{sec:black_box}).
Thus, there is a demand for new approaches that can tolerate 
larger distortion while still ensuring the generated adversarial examples being natural and sharing the same semantics of original images.

In this paper, we propose a \emph{ structure-preserving attack (SPA)} for generating natural
adversarial examples  with  
high transferability. Our approach is based on the hypothesis that the semantics of an image is mainly derived from its spatial structures~\cite{landau1988importance}. Thus, a promising approach to ensure semantic similarity is to maintain the core structural patterns across images. The main idea of our approach is to introduce \textit{structural perturbations} to images so that the generated adversarial images keep similar structures as the original images. 
Instead of giving an accurate definition of \textit{structures}, which is a challenging task, 
we adopt an intuitive definition of \textit{structure pattern} by partitioning pixels according to their intensity values (See Definition~\ref{def:structure_pattern}). 
In our SPA algorithm, we enforce that the same perturbation is applied to all the pixels
in the same structure pattern so that the computed perturbation for the given image is structural. 
By imposing the structural constraints, our approach can tolerate moderate to large distortion while still guaranteeing that the generated images are realistic and meaningful to human eyes.
Consequently, SPA allows relatively large distortion to the original images than typical small-perturbation-based attacks,
leading to 
better attack ability and transferability. This idea is further illustrated in Figure~\ref{fig:intro}, where structural and random perturbations are added to the original image, respectively, with the same maximum allowed distortion size of 20/255.
We observe that unlike random perturbation, the structural perturbation maintains the structure of the original image and is semantically meaningful to human beings. Thus, even under relatively larger distortion, the adversarial example generated through structural perturbation is still natural. In contrast, the random perturbation 
may destroy the structure of the original image and further dirty the image to some extent.
We show that our structure-preserving adversarial examples are highly transferable with little loss of successful attack rate when applied to black-box attacks even when a defense mechanism~\cite{madry2018towards,randomized_smoothing}
is applied. 

This work broadens the scope of adversarial machine learning by showing a new class of adversarial examples that follow different distributions from the training dataset while still being legible and natural to humans. Our study reveals the weakness of current defense mechanisms in the face of structure-preserving attacks that 
relax the
small perturbation constraint.

Our main contributions can be summarized as follows.
\begin{itemize}
      \item We propose the structure-preserving attack (SPA) as a new approach for generating natural adversarial examples with strong transferability. The proposed structural perturbation concept is a general idea and can be 
      combined with 
      small-perturbation-based attacks to generate even stronger attacks.

    \item 
    We conduct comprehensive experiments and
    demonstrate that SPA adversarial examples achieve extremely high transferability even when defense mechanisms are applied. 
    Further, when combined with other attacks, SPA strikingly enhances both the white-box and black-box attack abilities.



    \item We conduct adversarial training with SPA 
    and show that even SPA-based adversarial training hardly resist SPA itself, which further demonstrates 
the effectiveness of SPA.
  \item We analyze the relationship between attack ability and attack space from the perspective of \textit{space flexibility} and \textit{distortion flexibility}. We show 
  that to obtain strong attack ability, it is profitable 
  to sacrifice a bit of space flexibility in exchange for greater distortion flexibility. 



\end{itemize}

\section{BACKGROUND}
In this section, we briefly review recent studies on adversarial examples and defense mechanisms.

\subsection{Adversarial attacks}
Many adversarial attacks have been proposed~\cite{carlini2017towards,2014arXiv1412.6572G,madry2018towards,chen2017ead}, most of which are small-perturbation-based (measured by a $L_p$-norm for some $p$). 
Among them, the Projected Gradient Descent (PGD) method is the most effective $L_{\infty}$-norm based attack 
for naturally trained networks and has good transferability, while $L_{2}$-norm based CW attack is the most effective white-box attack for deterministic networks, 
including both naturally trained networks and PGD-based adversarially trained networks. A neural network is considered  deterministic if the model does not utilize any randomization.

\subsection{Defense techniques}

To mitigate the threat of adversarial examples, a number of defense mechanisms have been proposed in the literature~\cite{kurakin2016adversarial,madry2018towards,guo2018countering,cisse2017parseval}.
However, most defense methods are ineffective in the newly proposed attacks~\cite{athalye2018obfuscated,carlini2017adversarial}. Only a few state-of-the-art defense models have demonstrated their robustness to adversarial examples~\cite{madry2018towards,randomized_smoothing}. 
In particular, PGD-based adversarial training~\cite{madry2018towards} is an effective defense to resist $L_\infty$ attacks include PGD itself, while randomization has been shown to be an effective technique to resist the $L_2$-based CW attack as the randomness in target networks makes it difficult to compute accurate adversarial perturbations.
A representative randomization technique is randomized smoothing, where a Gaussian noise layer is sited in the front of  the target model~\cite{randomized_smoothing}.
 To shed light on the weakness of existing defense mechanisms, we 
will evaluate SPA against
PGD-based adversarial training and randomized smoothing. Details are given in the evaluation section.
\section{STRUCTURE-PRESERVING ATTACKS TO DNNS}\label{sec:method}


\begin{figure}[!t]
\vspace{-1mm}
\centering
\includegraphics[height=1.5  in,width=0.45 \textwidth]{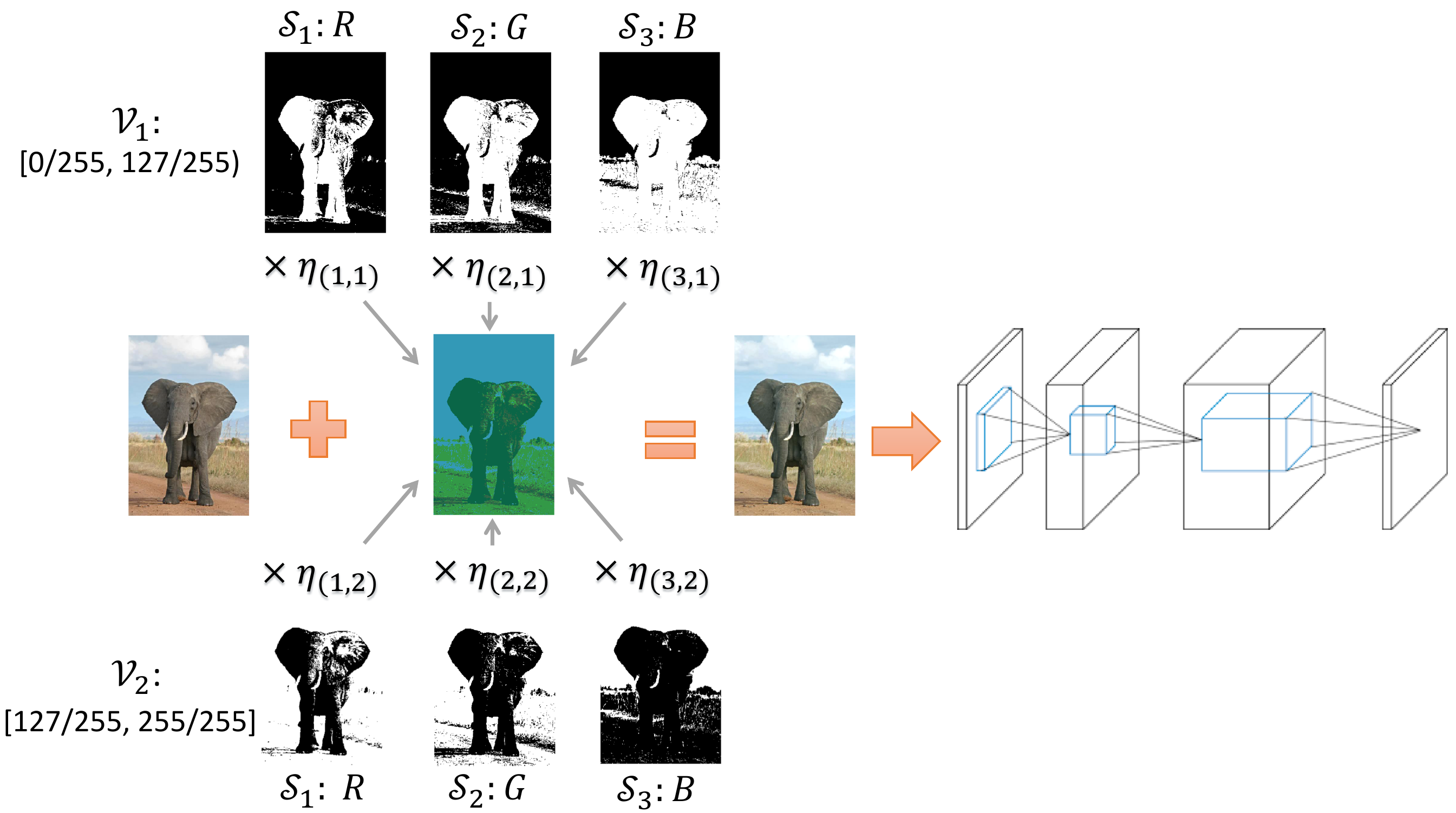}
\caption{The overall network architecture for generating SPA adversarial examples. The SPA layer sits in front of the target model.}
\label{fig:overall_network}
\end{figure}

The proposed approach is called \textbf{S}tructure-\textbf{P}reserving
\textbf{A}ttack (SPA), which attempts to generate 
structural adversarial perturbations
that maintain the same structure as the original images. Similar to previous small-perturbation-based attacks, our structural adversarial examples are generated by 
introducing perturbations to the original images. The key difference, however, is that instead of perturbing each pixel in an image independently as in most previous work, pixels~\footnote{By definition, a pixel in an image is the composition of all the channels in a specific position in the image. In this work, a pixel sometimes refers to the feature of a single channel in a specific position when there is no ambiguity.} in the same structure pattern (to be formally defined below) are perturbed similarly in our approach. By considering the intrinsic 
structure of images, 
structural adversarial examples generated under relatively large perturbations are comparably natural or even more natural than traditional small-perturbation-based adversarial examples (see our experiment results in the next section).
Thus, 
SPA 
exhibits stronger
transferability.

SPA can be either untargeted or targeted and can be used for both white-box and black-box attacks. In this paper, we focus on untargeted white-box and black-box attacks. More specifically, 
for a given image $\vect{x}$, we attempt to find a small perturbation $\vect{\epsilon}$
so that the adversarial perturbation $\vect{\epsilon}$ keeps the structure of the original image, 
while the prediction on  new image $\vect{x'} = \vect{x}+\vect{\epsilon}$ reported by the target model $f$
is different from the ground truth label $C_\mathrm{true}(\vect{x})$. Note that $C_\mathrm{true}(\vect{x})$ is the ground truth of the original image $\vect{x}$. However, since the adversarial example should be semantically similar to the original image, they share the same ground truth. Formally, we require that
\begin{eqnarray}\label{eq:structured_adv_example}
\argmax_{y \in \mathcal{Y}} f_y(\vect{x +\vect{\epsilon}}) \ne C_\mathrm{true}(\vect{x}), \ \ \|\vect{\epsilon}\|_p \leq \delta, \text{ and } \ \ \vect{\epsilon} \sim \vect{x}\end{eqnarray}
where 
we use `$\sim$' to denote 
the fact that the structures of two images are similar to each other. 
Notice that the first two conditions are commonly adopted in small-perturbation-based adversarial examples, while the last condition is unique to our approach. By explicitly introducing the structure-preserving perturbation constraint, our approach can tolerate relatively large perturbations while still keeping the generated adversarial examples natural.
The key idea that differentiates our approach from typical small-perturbation-based attacks is to ensure the perturbations being structural, which in turn implies structural adversarial examples. 
Although small-perturbation-based adversarial examples also keep the structure of original images, 
this is completely achieved through the small perturbation restriction. In contrast, 
structure-preserving is achieved through 
the structural perturbation restriction in our approach, which is orthogonal to 
the small perturbation restriction. This is the key reason why our approach 
may relax the small perturbation restriction. 





We then discuss our approach for measuring structure similarity. Intuitively, a structure in an image refers to a continuous region of pixels that constitute an object or the background aligned with human perception.
For example, we may consider the elephant, cloud, blue sky, road and meadow as structure patterns in Figure~\ref{fig:intro}. 
Different structures (objects) can be roughly identified based on pixel intensities 
as a structure (object) in an image often consists of pixels of similar colors. Therefore, although it is difficult to give an accurate definition of structures, we approximate a structure pattern with all pixels with similar pixel values and nearby, which is formally defined as follows:


\begin{definition} \label{def:structure_pattern}
\textbf{Structure Pattern:} 
Let $\mathcal{S}$ be a space partition that divides the set $[1,...,W] \times [1,...,H] \times [1,...,C]$ into disjoint sub-regions. For the $s$-th sub-region in $\mathcal{S}$, let $\mathcal{V}_s$ denote a partition that divides pixel value range
into disjoint sub-intervals. 
A structure pattern of indices $(s,v)$ is defined as the set of pixels for which the pixel position lies in the $s$-th sub-region in $\mathcal{S}$ and the pixel value lies in the $v$-th sub-interval in $\mathcal{V}_s$. 
For simplicity, we assume $|\mathcal{V}_s|$ is the same for all $s$, and let $S = |\mathcal{S}|$ and $V = |\mathcal{V}_s|$.

\end{definition}


According to this definition, a structure pattern consists of a set of pixels that share similar pixel values and are in close proximity. It is worth noting that in traditional $L_p$-norm based perturbations, each pixel is treated {\it independently}, which can be viewed as a specific case of our definition, where each structure pattern has a single pixel.
Note that Definition \ref{def:structure_pattern} provides a general definition of structure patterns and applies to any possible partitions $\mathcal{S}$ and $\{\mathcal{V}_s\}$. In our experiment discussed below, the whole space is partitioned into three sub-regions corresponding to the three channels (i.e., $S=3$). Further, an even partition of pixel values is applied to all the channels. Figure~\ref{fig:overall_network} gives an example of this approach where $S = 3$ and $V = 2$. This simple approach for generating partitions already provides satisfactory results. However, our approach discussed below applies to any given partitions.


Our main idea for keeping structure patterns unchanged under perturbations 
is to guarantee that all the pixels in the same structure pattern are perturbed by a {\it similar} amount. That is, instead of perturbing each pixel independently as in most previous work, we consider structural perturbations that are aligned with the structure patterns of images. To simplify the implementation, we consider a special case in this work by requiring that all the pixels in the same structure pattern are perturbed by the {\it same} amount.

Formally, we define a {\it meta-perturbation} as an $S \times V$ matrix $\vect{\eta}$ where $\vect{\eta}(s,v)$ gives the perturbation to be added to the pixels in the structure pattern of indices $(s,v)$. Let $\vect{\epsilon} =  \texttt{pert}(\vect{x},\vect{\eta})$, where \texttt{pert} is a function that generates the perturbation $\vect{\epsilon}$ for image $\vect{x}$ according to the meta-perturbation $\vect{\eta}$. An implementation of \texttt{pert} is given in Algorithm~\ref{algor:spa} (lines 9-19), where for each pixel, its structure indices in the space and value partitions are first identified and the corresponding meta-perturbation value is then obtained. {Algorithm~\ref{algor:spa} can be made more efficient through vectorization. We choose the current form to illustrate the main idea more clearly.} Alternatively, for each structure pattern of indices $(s,v)$ in an image $\vect{x}$, we may define a binary mask (a black-white image of the same shape as image $\vect{x}$) $\vect{b}_{s,v}$, where 
$\vect{b}_{s,v}(m,n,c) = 1$
if the pixel $(m,n,c)$ is in the structure pattern $(s,v)$ and $\vect{b}_{s,v}(m,n,c) = 0$ otherwise. Figure~\ref{fig:overall_network} shows the six binary masks for the original image $\vect{x}$ on the left. The desired perturbation for $\vect{x}$ can then be found by taking a weighted sum of the binary masks with the weights taken from $\vect{\eta}$. That is, $\texttt{pert}(\vect{x},\vect{\eta}) = \sum_{s,v}\vect{\eta}(s,v)\vect{b}_{s,v}$.

Our objective is to find a meta-perturbation $\vect{\eta}$ that meets the following condition:
\begin{align}
\argmax_{y \in \mathcal{Y}} f_y(\vect{x} + \texttt{pert}(\vect{x},\vect{\eta})) & \ne C_\mathrm{true}(\vect{x}), \nonumber \\
\|\texttt{pert}(\vect{x},\vect{\eta})\|_p & \leq \delta
\end{align}

\noindent 
It is important to mention that
in addition to the misclassification goal and the small perturbation restriction, we further require that the generated perturbation preserves the structure patterns of the original images, which is achieved by the \texttt{pert} function. Also note that $ \|\texttt{pert}(\vect{x},\vect{\eta})\|_p = \|\vect{\eta}\|_p$ for $p=\infty$.
\subsection{Generating structural adversarial perturbations}
SPA is trained together with the target model by adding an extra layer in front of the target network (see Figure~\ref{fig:overall_network}).
For untargeted attacks, our objective is to find a small meta-perturbation $\vect{\eta}$ so that $\argmax_{y \in \mathcal{Y}}{f_y(\vect{x} +\texttt{pert}(\vect{\eta},\vect{x})})\ne C_\mathrm{true}(\vect{x})$ for any $\vect{x} \in \mathcal{X}$.
To improve the chance of successful attacks, we aim to find $\vect{\eta}$ so that the distance between the predicted logits
and the ground truth is maximized for a given image.
Let $\vect{l}(\vect{x})$ be a $|\mathcal{Y}|$ dimensional vector where $l_y(\vect{x})=1$ if $y = C_\mathrm{true}(\vect{x})$ and $l_y(\vect{x})=0$ otherwise.
We then solve the following optimization problem to find the  meta-perturbation $\vect{\eta}$ for a given image $\vect{x}$:
\begin{equation}\label{eq:obj}
\begin{aligned}
\underset{\vect{\eta}}{\mathrm{argmax}} \ \ 
\mathcal{J}(f(\vect{x} +\texttt{pert}(\vect{x},\vect{\eta})),\vect{l}(\vect{x})) \\
s.t. \ \ \|\texttt{pert}(\vect{x},\vect{\eta})\|_p \leq \delta
\end{aligned}
\end{equation}

\noindent where
$\mathcal{J}$
 is a loss function that measures the difference between the output logit $f(\vect{x} + \texttt{pert}(\vect{\eta}, \vect{x}))$ when the target model $f$ is applied to the crafted adversarial example $\vect{x} + \texttt{pert}(\vect{\eta}, \vect{x})$ and
the ground truth of the original image. In this work, we use the cross entropy as the loss function.

As SPA sits in front of the target network as shown in Figure \ref{fig:overall_network}, we fix the target model when solving the optimization problem~\eqref{eq:obj} to generate SPA adversarial examples. 
Projected gradient
descent (PGD) is a standard technique to solve $L_{p}$-constrained optimization problems~\cite{Bahmani_2013}. It has recently been 
used to design adversarial attacks and PGD attack has become a benchmark attack~\cite{madry2018towards}~\footnote{In this work, PGD refers to PGD attack unless otherwise specified.}. 
In this paper, we use PGD to solve the above constrained optimization problem for finding the optimal parameters $\vect{\eta}$ and adopt $L_{\infty}$ as the norm metric as in PGD attack. We highlight that our SPA approach is a general idea and can be applied to most existing attacks~\cite{2014arXiv1412.6572G,carlini2017towards,chen2017ead}~\footnote{Even though we consider SPA as a constrained optimization problem in the paper, SPA can also be formulated as an unconstrained problem with the $L_{p}$ norm restriction included in the objective function as CW~\cite{carlini2017towards} and EAD~\cite{chen2017ead} attacks.}.

The sign value of the gradient multiplied by a constant step size $\sigma$ is then used to update $\vect{\eta}$ (line 6).  
 We then project $\vect{\eta}$ to satisfy the small perturbation constraint. For $L_{\infty}$-norm based perturbation constraint, this can be easily implemented using the $\texttt{clip}$ function to restrict the perturbation to fall into the maximum allowed distortion range $[-\delta, \delta]$ (line 7). 
We highlight that instead of computing the gradient with respect to the perturbation itself as in standard PGD attack~\cite{madry2018towards}, we compute the gradient with respect to the meta-perturbation  
and then use the computed meta-perturbation to form the structural perturbation. This is crucial for ensuring the perturbation generated by SPA being structural and is the main difference between SPA and PGD.  

Due to the good convergence property of the cross entropy loss function and the small number of parameters ($S \times V$ parameters) to be optimized,
the searching for the optimal meta-perturbation $\vect{\eta}$ converges within a small number of iterations.

\begin{algorithm}[tb]
\SetAlgoLined
\DontPrintSemicolon
  \KwInput{target model $f$; original labelled images
  ($\vect{x}^{[W, H, C]}, \vect{l}(\vect{x})$);
    a space partition $\mathcal{S}$ and a set of pixel value partitions $\{\mathcal{V}_s\}$ where $|\mathcal{S}|=S$ and $|\mathcal{V}_s|=V$ for any $s$; 
  maximum perturbation size $\delta$;
  step size $\sigma$; number of steps $K$}

  \KwOutput{meta-perturbation $\vect{\eta}$}



  %


  
  $\vect{\eta}^{(1)} = \vect{0}^{[S,V]}$;\\
  \For {$k \in [1, ...,K]$}
  {
  $\vect{\epsilon} =$ \texttt{pert}($\vect{x},\vect{\eta}^{(k)}$); \\
  $L = \mathcal{J}(f(\vect{x} + \vect{\epsilon}),\vect{l}(\vect{x}))$; \\
  $\bigtriangledown_{\vect{\eta}^{(k)}}L  = \frac{\partial L}{\partial \vect{\vect{\eta}}}\mid_{\vect{\eta}^{(k)}}$;\\
$\vect{\eta}^{(k+1)} = \vect{\eta}^{(k)} + \sigma \times sign(\bigtriangledown_{\vect{\eta}^{(k)}}L )$;\\
  \tcp{project each entry in $\vect{\eta}$ into $[-\delta, \delta]$}
  $\vect{\eta}^{(k+1)} =$ \texttt{clip}$(\vect{\eta}^{(k+1)}, -\delta,\delta)$;\\
  }


\SetKwFunction{FMain}{pert}
\SetKwProg{Fn}{def}{:}{}
\Fn{\FMain{$\vect{x}$, $\vect{\eta}$}}{

\For {$c \in [1, ...,C]$}{
\For {$m \in [1, ...,W]$}{
\For {$n \in [1, ...,H]$}{
    \tcp{find the sub-region in $\mathcal{S}$ where the pixel is in}
    $s = \texttt{space\_index}(m,n,c,\mathcal{S})$; \\
    \tcp{find the sub-interval in $\mathcal{V}_s$ where the pixel is in}
    $v = \texttt{value\_index}(\vect{x}(m,n,c),\mathcal{V}_s)$;\\
    $\vect{\epsilon}(m,n,c) = \vect{\eta}(s,v)$;\\
}
}
}
\textbf{return} $\vect{\epsilon}$
}

\caption{Structure-Preserving Attack against DNNs}
\label{algor:spa}
\end{algorithm}

\section{EXPERIMENT RESULTS}\label{sec:exp}





To evaluate the performance of SPA, we compare it with two baseline attack algorithms,
PGD attack~\cite{madry2018towards} and CW attack~\cite{carlini2017towards} on two popular 
image classification datasets,
MNIST~\cite{lecun1998gradient},
 and CIFAR10~\cite{cifar}.
Moreover, the attack methods are evaluated both on the vanilla models with different architectures and when the two state-of-the-art defense mechanisms, PGD-based adversarial training~\cite{madry2018towards} and randomized smoothing~\cite{randomized_smoothing}, are applied.


\subsection{Experiment settings}\label{sec:setup}



\smallskip
\noindent{\bf Evaluation metrics:}
As in most previous works, we report the classification accuracy of target models under various attack-defense configurations. For $L_2$-based CW attack, 
both the accuracy and distortion size 
determine its attack ability. Thus, we also report the $L_{2}$ distortion size for CW attack.

A stronger attack method leads to a lower classification accuracy while a target model with stronger defense ability has a higher accuracy. Evaluation is conducted in both white-box and black-box attack settings~\cite{2018arXiv180100553A}.

\smallskip 
\noindent{\bf Datasets:}\label{sec:datasets}
We use two popular image classification datasets,
MNIST~\cite{lecun1998gradient}, and CIFAR10~\cite{cifar}. 
The pixel value of MNIST is a real number  in [0, 1], and the pixel value of CFIAR10 is an integer in [0, 255]. 


\smallskip
\noindent{\bf Baseline attacks:}\label{sec:baseline_attack}
For the two baseline attacks, CW is implemented using the source code~\footnote{\url{https://github.com/carlini/nn_robust_attacks}} in~\cite{carlini2017towards}
with the default configurations 
except that 1000 attack iterations are used to obtain stronger attack ability. PGD is implemented using source code~\footnote{\label{note_adv_tr_free}\url{https://github.com/ashafahi/free_adv_train}} in~\cite{shafahi2019adversarial} with the default settings. To enhance attack ability, we further combine SPA with the two baseline attacks and evaluate their effectiveness.

\smallskip
\noindent{\bf Baseline defenses:}
We evaluate SPA against two known defense techniques, PGD-based adversarial training~\cite{madry2018towards} and randomized smoothing~\cite{randomized_smoothing}. PGD-based adversarial training on CIFAR10~\footref{note_adv_tr_free}and MNIST~\footref{madry_mnist_challenge} is implemented with default parameters.
PGD-based adversarial training has been demonstrated to be the most effective defense method against $L_{\infty}$-norm based attacks on the MNIST and CIFAR10 datasets~\cite{athalye2018obfuscated}. It can also be generalized to defend other $L_{p}$-norm attacks. Randomized smoothing 
is the most practically effective method to defend $L_{2}$-norm based attacks~\cite{araujo2019robust}.

\smallskip
\noindent{\bf Target models:}
For MNIST, we use two variants of LeNet~\cite{lecun1998gradient}, namely \textit{LeNet1} and \textit{LeNet2}. LeNet1 is regarded as the primary network for MNIST in our experiments.
For CIFAR10, we adopt WideResNet-32$\times$10~\cite{zagoruyko2016wide} as the primary network and also use ResNet-32~\cite{he2016deep} as a different network architecture for comparison.
Both LeNet1 and  WideResNet-32$\times$10 have been used 
in the Madry's MNIST~\footnote{\label{madry_mnist_challenge}\url{https://github.com/MadryLab/mnist_challenge}} and CIFAR10 Adversarial Examples Challenges~\footnote{\url{https://github.com/MadryLab/cifar10_challenge}
}, respectively.
In each case, these models are trained  both under naturally and adversarially setting. 


To train a randomized smoothing model, we add a Gaussian noise layer with zero mean 
in front of the original target model and retrain it to get the randomized smoothing model.
The standard deviation of the Gaussian noise is set to 1.0 and 64 (0.25$\times$255) for MNIST and CIFAR10, respectively, following the settings in~\cite{randomized_smoothing}.


\smallskip
\noindent{\bf Parameter settings in SPA:}
For the MNIST 
dataset, 
we simply partition the pixel value range $[0,1]$ into 255 equal-sized intervals, which is aligned with the standard image quantification and representation methods. For SPA based white-box attack, we set the maximum allowed perturbation size $\delta$ to 0.4, which is larger than the default value of 0.3 in PGD. The step size $\sigma$ is set to 0.01 and the number of steps $K$ is set to 40 (same as the default setting in PGD) in Algorithm~\ref{algor:spa}. 
For SPA+PGD attack, we first generate a SPA adversarial image with $\delta=0.4$ and then apply PGD attack with the default PGD setting except using a smaller perturbation size $\delta=0.1$ to ensure that the generated image is natural. For SPA+CW attack, we follow the same SPA setting and the aforementioned CW setting.  

For the CIFAR10
dataset, each image is partitioned into 3 sub-areas corresponding to the three channels (RGB). For each channel, the pixel value range is then evenly partitioned into 255 intervals, similar to the MNIST dataset.
 We set the maximum allowed perturbation size $\delta$ to 20, which is larger than the default value 8 used in PGD attack. We will show that despite the large difference in the perturbation size, the generated SPA images and PGD images are comparable in terms of how natural they are when viewed by human beings. 
In Algorithm~\ref{algor:spa}, the step size $\sigma$ is set to 2 and the number of steps $K$ is set to 20 (both follow the same 
default setting in PGD attack). For SPA+PGD attack, we first generate an SPA adversarial image and then apply PGD attack with a smaller perturbation size $\delta$ = 2. 
Again, we use the same SPA and CW setting for the SPA+CW attack.


It should be pointed out that
one-shot attacks are inefficient for models protected by randomized transformations to the input as in the case of randomized smoothing models~\cite{athalye2018obfuscated}. In this case, we adopt the approach of Expectation over Transformation (EOT) in \cite{athalye2017synthesizing} and compute the gradient over the expected transformation. We set the number of EOT to 50, following the setting in~\cite{araujo2019robust}.



\subsection{Evaluation results for white-box attacks}\label{sec:white_attack}

\begin{figure}
\centering
\includegraphics[height=1.90 in,width= 0.50  \textwidth]{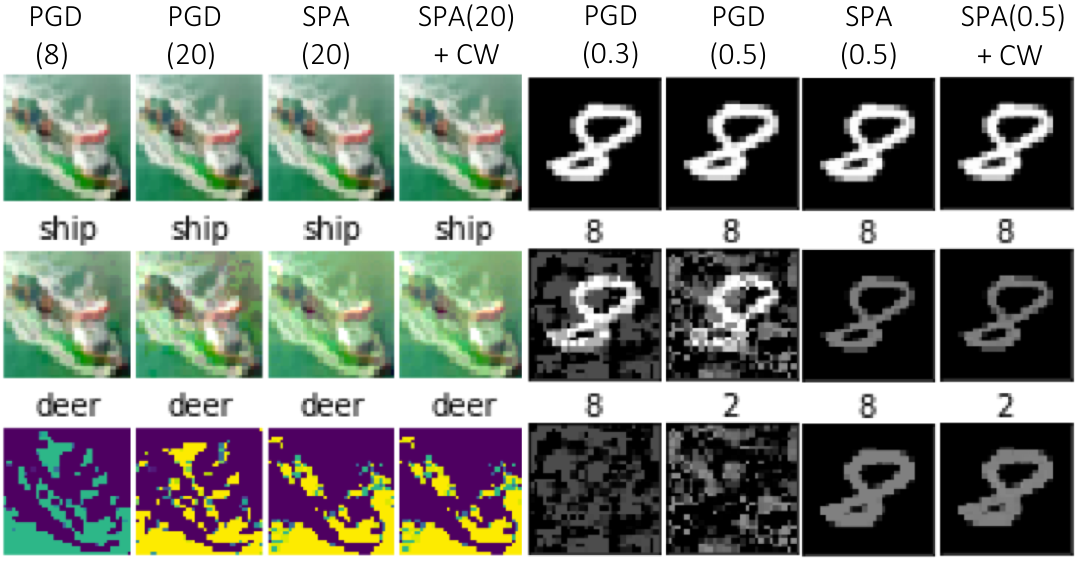}
\caption{
Examples of adversarial images generated by SPA, PGD and SPA+CW attacks. The three rows show the original images, the adversarial images and the corresponding perturbations, respectively, generated by the aforementioned attack models against the PGD-based adversarially trained primary networks on the CIFAR10 (left) and MNIST (right) dataset.
The maximum allowed $L_\infty$ perturbation sizes $\delta$ are shown in parentheses.
\label{fig:illustration_adv_examples}}
\vspace{-1mm}
\end{figure}

\begin{figure*}
\vspace{-6mm}
\centering
\includegraphics[height=1.70  in,width=0.45\textwidth]{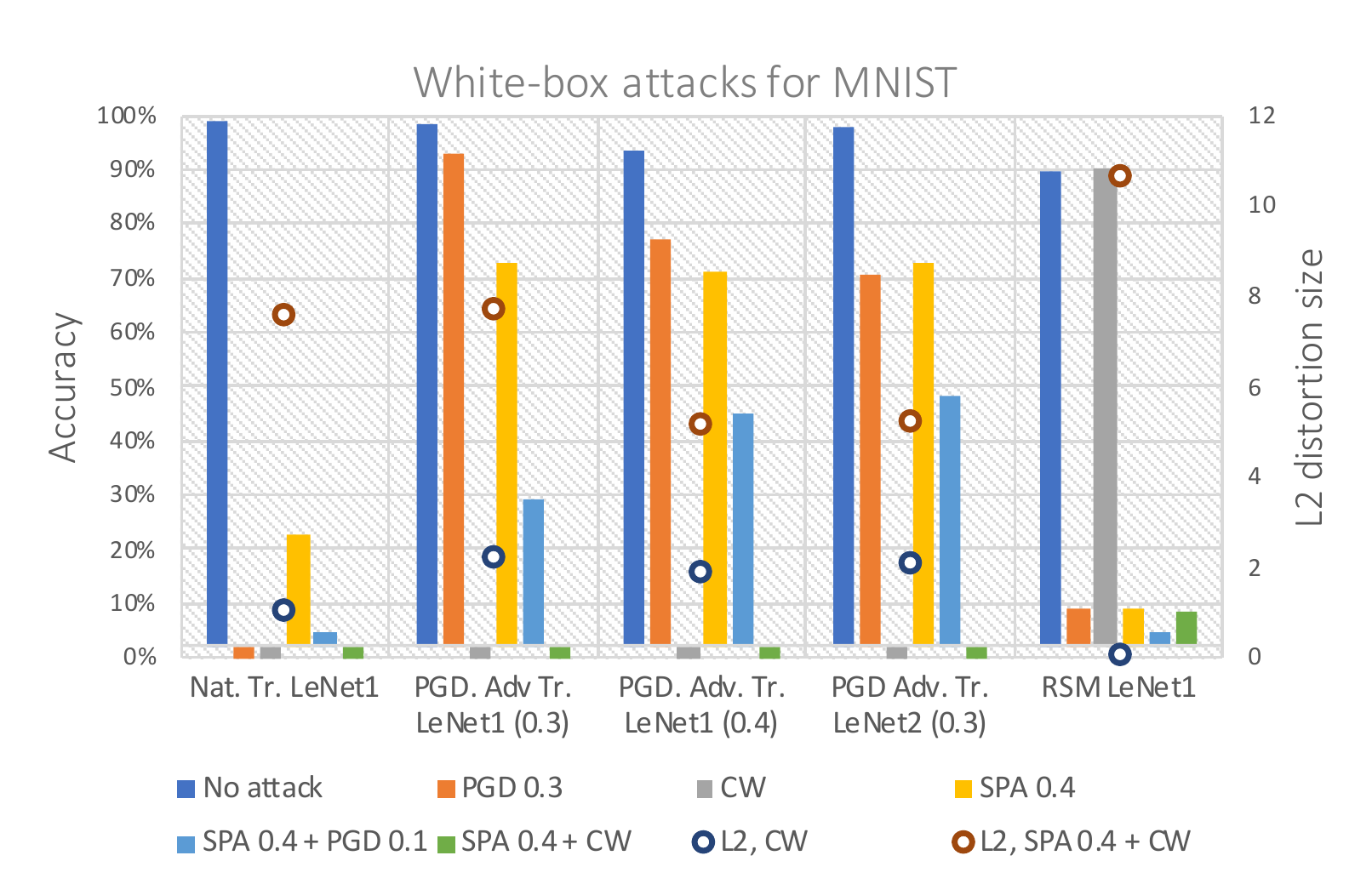}
\hspace{5mm}
\includegraphics[height=1.70  in,width=0.45\textwidth]{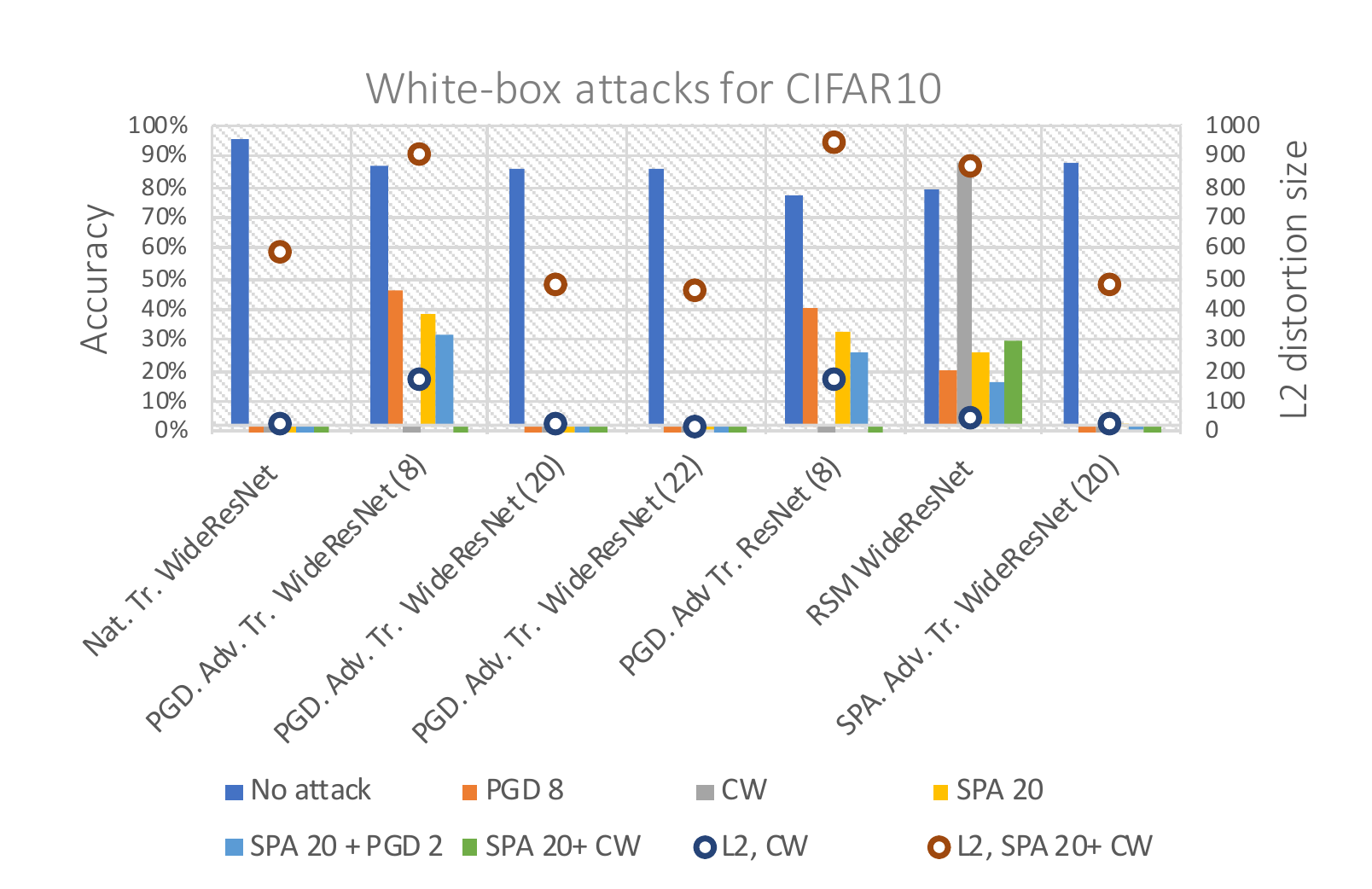}

\caption{\textbf{White-box attacks}. Classification accuracy of the target networks under different attack-defense configurations on the MNIST (left) and CIFAR10 (right) datasets. 
 ``Nat. Tr.'' denotes the naturally trained network, ``Adv. Tr'' denotes the adversarially trained network and ``RSM'' represents the RSM model. The distortion sizes used in adversarial training are shown in parentheses.}
\label{fig:white_attack}
\end{figure*}

We report the evaluation results for white-box attacks in Figure~\ref{fig:white_attack}. For PGD-based adversarial training, 
we show the results for both standard PGD distortion sizes (0.3 for MNIST and 8 for CIFAR10) and larger sizes (0.4 for MNIST and  20 also 22 for CIFAR10) for fair comparison. We note that when the LeNet1 model is PGD-adversarially trained with a distortion size beyond 0.5, the accuracy (on clean images) drops below 10\%. We consider the reason is that the large unstructured perturbations have destroyed the structure of the images. Thus we do not report the results for more larger distortion size for PGD-based adversarial training.


It is observed that the performance of SPA+CW is the best against nearly all target models under white-box attack setting. CW is good at attacking deterministic networks including both naturally trained and adversarially trained networks. We note that CW is poor at attacking stochastic networks such as RSM. However, after integrating SPA, SPA+CW could ferociously attack the RSM.


We observe that 
SPA is superior to PGD when attacking PGD-based adversarially trained models as it allows a larger maximum perturbation size $\delta$. Further, SPA is 
comparable to PGD for both naturally trained models and RSM. 
Moreover, by simply combining SPA with PGD, the revised model demonstrates much better attack ability than the original models.
Thus, although the original SPA may not perform better than PGD and CW for all cases, it can be easily integrated with other attacks to obtain much better attack ability.

\subsection{Black-box attacks and transferability}\label{sec:black_box}

\begin{figure*}[!t]
\centering
\includegraphics[height=1.70  in,width=0.45\textwidth]{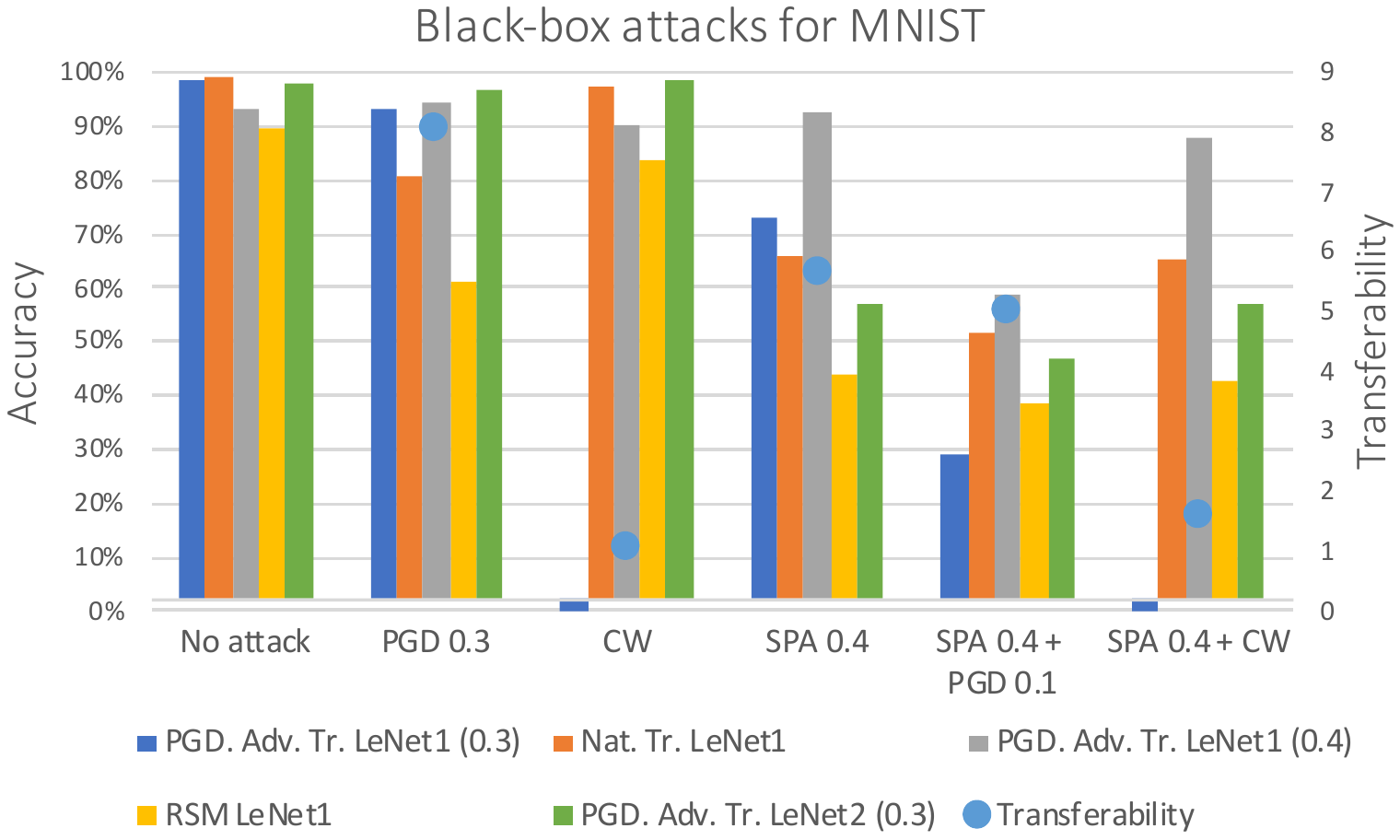}
\hspace{5mm}
\includegraphics[height=1.70  in,width=0.45\textwidth]{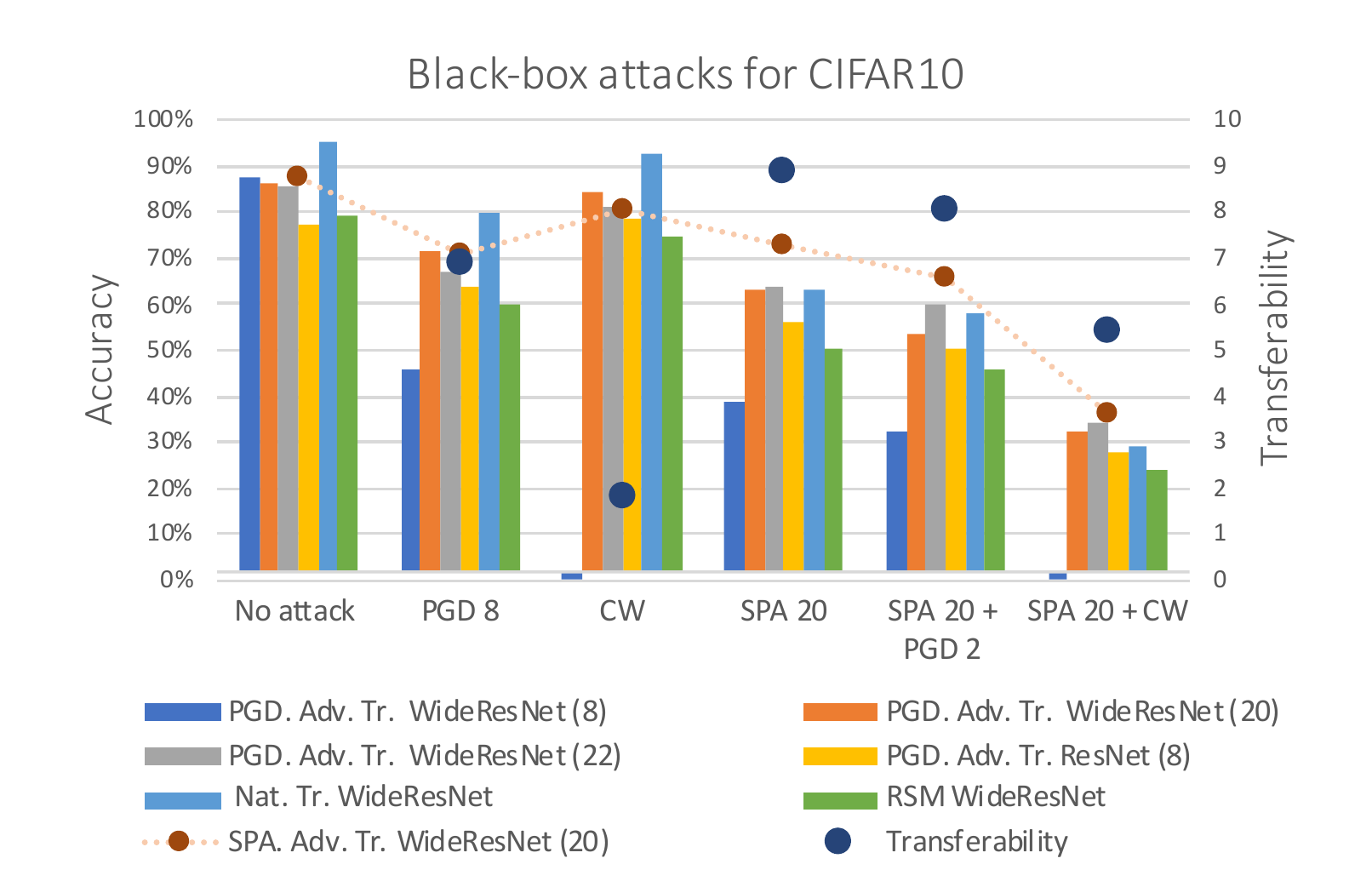}

\caption{\textbf{Black-box attacks}. Classification accuracy
on the MNIST (left) and CIFAR10 (right) datasets where the PGD-based adversarially trained primary
networks LeNet1 and WideNetwork-32$\times$10 are used as the substitute model for MNIST and CIFAR10, respectively.  Other models are used as target models. The distortion sizes used in adversarial training are shown in parentheses.}
\label{fig:black_attack}
\end{figure*}

In this section, we present experimental results for black-box attacks (reported in Figure~\ref{fig:black_attack}). In standard
transfer-based
black-box attacks (in contrast to query-based attacks~\cite{chen2017zoo,bhagoji2018practical}), attackers first specify a substitute model to the black-box model,
then generate a set of adversarial examples that could successfully attack the substitute model~\cite{papernot2016transferability,papernot2017practical,liu2016delving}. These generated adversarial examples are considered to have strong transferable attack ability and are consequently used to attack the target black-box model. 
For transfer-based black-box attacks,
their effectiveness relies on 
how easily adversarial samples produced to mislead a specific model can also mislead other models~\cite{papernot2016transferability}. 
Thus, the black-box attack ability in this case is determined by both white-box attack ability and transferability, where the latter depicts the accuracy consistency when images are tested with different models.
 To better analyze how the two factors influence black-box attack ability respectively, we disentangle them and define transferability as the reciprocal of the average absolute difference between the accuracy of a substitute model and that of a target model, where the average is taken over multiple target models for a fixed substitute model.
 

In order to demonstrate the black-box attack ability
 of SPA, 
we perform transfer-based black-box attacks across different target models evaluated on two datasets. In particular, 
the primary network for each dataset is used as the substitute model, and the rest networks are used as the target models (please refer to Section~\ref{sec:setup} for more details on network architectures).
 
%
%

 From Figure~\ref{fig:black_attack}, 
we observe that the transferability of SPA is
generally
higher than that of other attacks.
In particular, the transferability of SPA on CIFAR10 is 28.50$\%$ higher than PGD, and 386.75$\%$ higher than CW.
SPA consistently achieves low accuracy with or without defense
and is extremely effective in the black-box setting. Although SPA does not perform significantly better than others 
against adversarially trained LeNet1 (with distortion size 0.4), the performance of SPA+PGD is satisfactory. Furthermore, SPA with a larger distortion size performs even better. Note that the black-box attack accuracy decreases to 54$\%$ when the distortion size increases to 0.5, while the distorted images can still be recognized very well as shown in Figure~\ref{fig:illustration_adv_examples}. 
 
 We notice that although PGD adversarial examples also exhibit 
high transferability, they are not as satisfactory as SPA in the black-box attack setting due to their relatively weak white-box attack performance on the substitute model.

Similar to white-box attacks, SPA+PGD  exhibits better black-box attack ability than the original PGD and SPA.
On the other hand, CW adversarial examples have demonstrated poor transferability and black-box attack ability in all the scenarios as shown in Figure~\ref{fig:black_attack}. 
However, when combined with SPA, the black-box attack ability of SPA+CW improves significantly 
compared with CW itself.
In particular, 
SPA+CW clearly outperforms 
all other attacks on CIFAR10. 

%
%

\subsection{The effect of attack space on attack ability}

\begin{figure*}
\vspace{-5mm}
\centering
\includegraphics[height=1.70  in,width=0.45\textwidth]{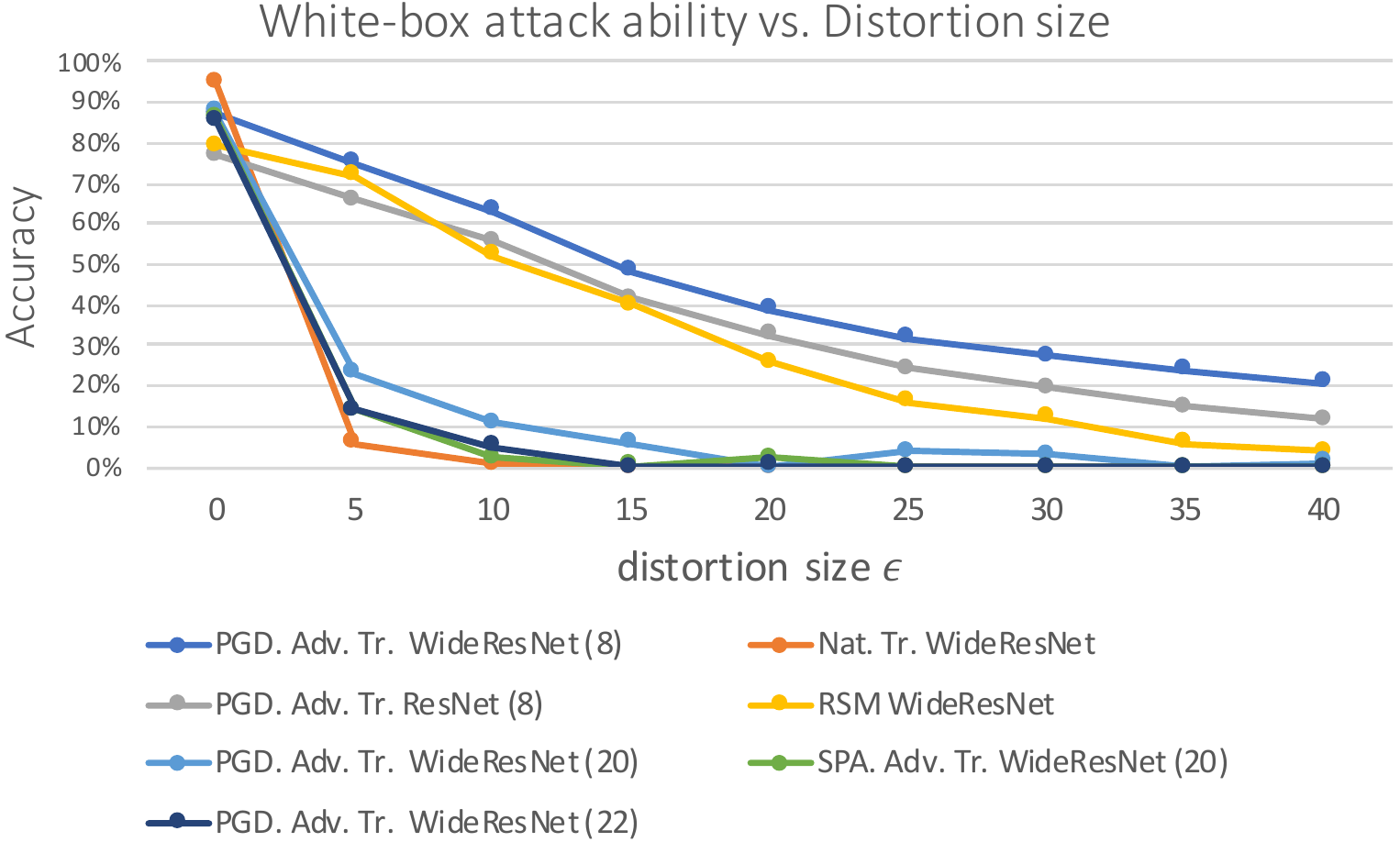}
\hspace{5mm}
\includegraphics[height=1.70  in,width=0.45\textwidth]{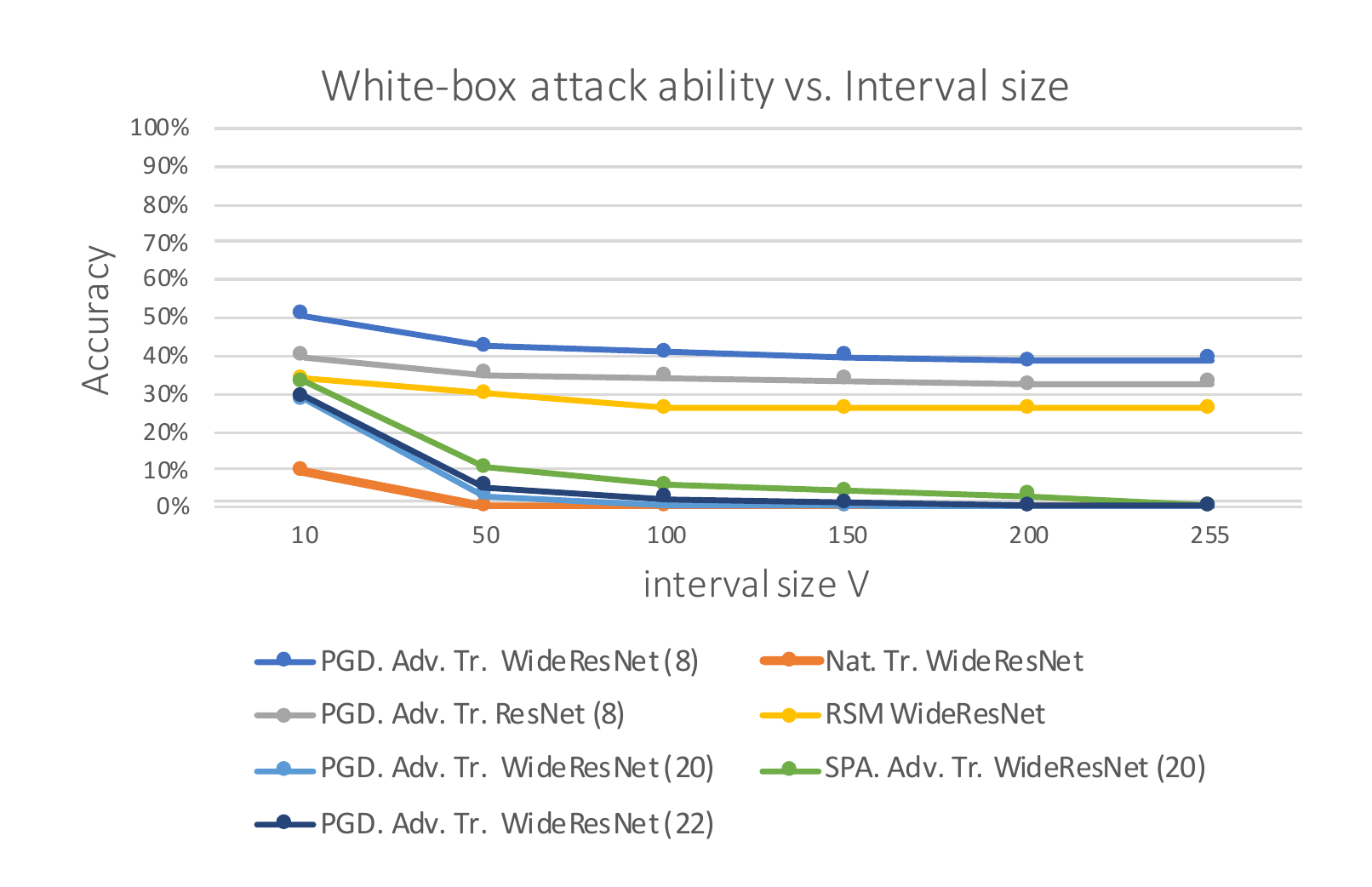}

\caption{The effect of distortion size and interval size on white-box attack ability.
With the increase of distortion size $\epsilon$,  the white-box attack ability consistently
increases. However, enlarging interval size V only 
improves 
white-box attack ability marginally after a certain threshold.}
\label{fig:attack_ability}
\end{figure*}

\begin{figure*}[!t]
\centering
\includegraphics[height=1.70  in,width=0.45\textwidth]{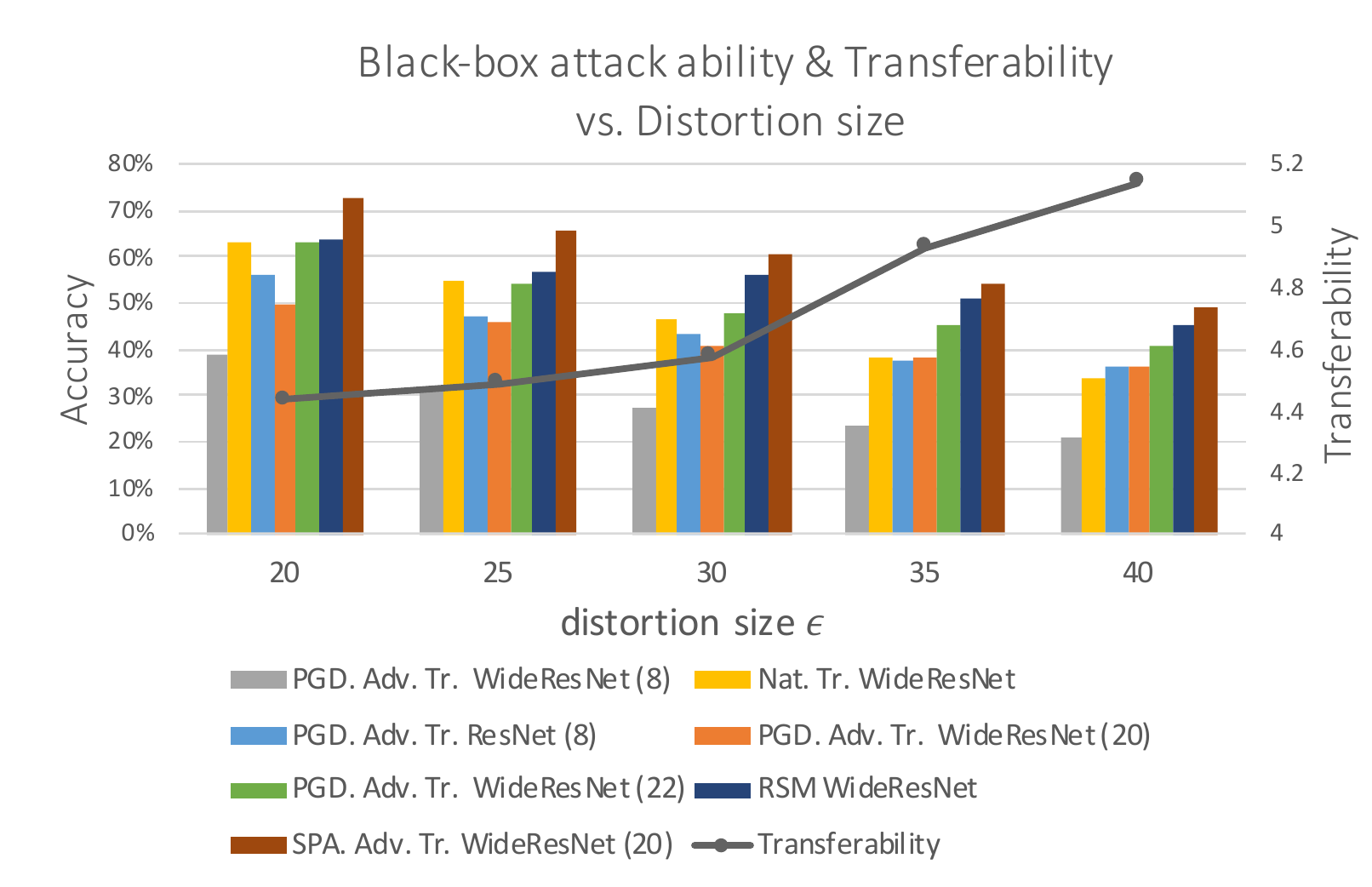}
\hspace{5mm}
\includegraphics[height=1.70 in,width=0.45\textwidth]{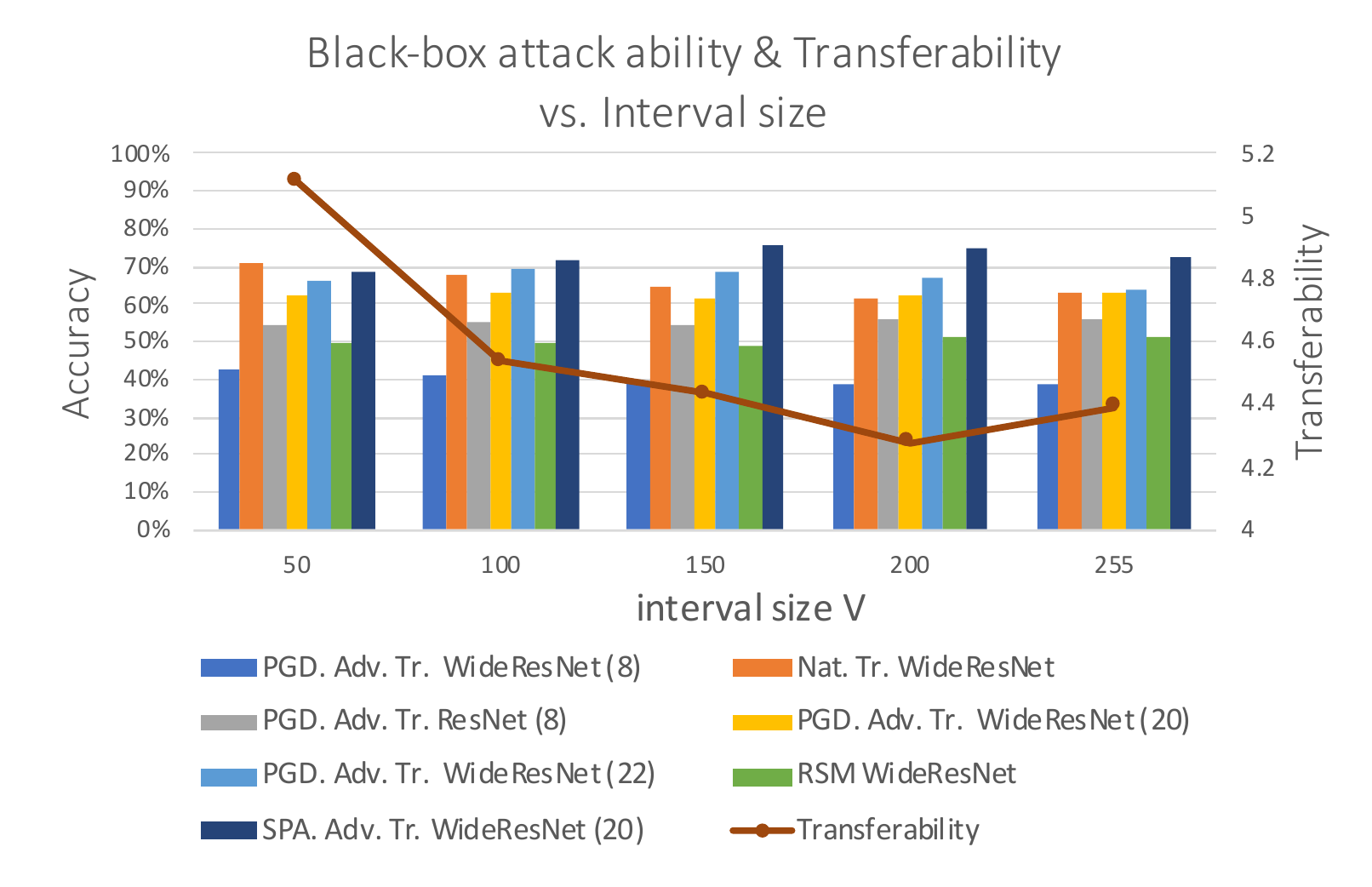}
\caption{The effect of distortion size and interval size on black-box attack ability. Similar to white-box attacks, increasing the distortion size $\epsilon$ profoundly raises the transferability and black-box attack ability.
Nevertheless, increasing the interval size $V$ may reduce the transferability; hence it has limited effect on the black-box attack ability. 
}
\label{fig:transferability}
\end{figure*}

To simplify the discussion below, we first define two terms, namely, the \textit{space flexibility} and the \textit{distortion flexibility} of the attack space. Intuitively, 
the attack space denotes 
the flexibility that the attacker has in modifying images, which can be measured from two dimensions. The \textit{space flexibility} refers to how many pixels in an image can be altered \textit{independently} by the attacker, while the \textit{distortion flexibility} measures to what extent each pixel can be modified (that is, the amount of perturbation that can be applied to each pixel). 
Traditional attacks independently twist each pixel, thus the space flexibility is $W \times H \times C$, where $W$, $H$ and $C$ are the width, height and the number of channels of an image, respectively.
 For our proposed structure based perturbation, all the pixels in a structure pattern are altered by the same amount, thus the number of pixels that can be changed independently (space flexibility) equals to the number of structure patterns $S \times V$, which is far smaller than that of the traditional attacks. On the other hand, our approach has greater distortion flexibility.


To understand how the \textit{space flexibility} and \textit{distortion flexibility} could affect the white- and black-box attack ability, we evaluate the performance of SPA 
by varying the distortion size $\epsilon$ and the interval size $V$ (we fix S to 3 in this experiment). 
The 
results are plotted in Figures~\ref{fig:attack_ability} and \ref{fig:transferability}. 
For white-box attack (see Figure~\ref{fig:attack_ability}), the distortion size is superior to the interval size in affecting white-box attack ability, and larger distortion confers quite higher white-box attack ability. For black-box attacks (see Figure~\ref{fig:transferability}), the transferability and black-box attack ability consistently improve with the increase of distortion size $\epsilon$. This 
confirms our motivational hypothesis that relaxing the small-perturbation restriction confers  higher transferability. 
However, there is a decreasing trend of transferability as the interval size $V$ enlarges. Thus it has little effect on the black-box attack ability. 


From these observations, we can infer that the distortion flexibility, rather than the space flexibility, plays a more important role in affecting white- and black-box attack ability. This is consistent with the
{\it i.i.d.} assumption that traditional supervised learning models including DNNs all rely on,  
where models become much less effective under distribution-shift data compared to testing data follow the same distribution as training data.
A large distortion size usually shifts data to  
a different data distribution, which invalidates target models on the distribution-shifted data.
Therefore, it is desired to moderately sacrifice space flexibility to allow for more distortion flexibility with the purpose of achieving higher black- and white- box attack ability simultaneously. This is the core contribution of our 
SPA approach. 

\subsection{Illustration of adversarial examples}\label{sec:illustration_adv_images}
Figures~\ref{fig:illustration_adv_examples} shows the adversarial examples generated by SPA, PGD and SPA+CW attacks (CW and SPA+PGD adversarial images are not shown due to space limitation). We observe that SPA adversarial examples indeed keep the structures of the original images. Although a larger maximum distortion is allowed than PGD, 
SPA adversarial examples are still clean and legible to humans. In particular, we observe that the SPA  adversarial examples on MNIST are far more natural to human eyes than the PGD adversarial examples generated under a smaller perturbation size. 
Moreover, it is interesting to observe that the generated images remain natural even when SPA is combined with  CW or PGD.

\subsection{Adversarial training with SPA}\label{sec:spa_adv_training}

Given the fact that adversarial training with PGD is less effective for SPA-based attacks, it is natural to wonder whether adversarial training with SPA is more effective. As SPA has far fewer parameters, it is reasonable to perform adversarial training with SPA. To this end, we conduct SPA-based adversarial training for the network WideResNet-32$\times$10 on the CIFAR10 dataset and compare it with PGD-based adversarial training. We stress that
it is extremely time-consuming to perform adversarial training by following the vanilla PGD-adversarial training paradigm. 
Inspired by the fast PGD-adversarial training proposed in~\cite{shafahi2019adversarial}, 
we design a fast SPA-adversarial training method 
by simultaneously computing the gradient with respect to the network weights and meta-perturbation $\vect{\eta}$. 
We train WideResNet-32$\times$10  using the following parameters: 80000 iterations, the batch-size is 128,  the replay parameter is 4, and the perturbation size 20. From the results in Figures~\ref{fig:white_attack} and~\ref{fig:black_attack}, 
we observe that SPA-based adversarial training does not achieve significant performance improvement for defending  
against SPA and other attacks even when compared with naturally trained models under white-box attacks. However, the SPA-based adversarially trained models have considerable black-box defense ability against SPA and the other two baseline attacks. Even so, SPA+CW can still satisfactorily attack the SPA-based adversarially trained model.


From the above results, we conclude SPA and its variants are 
profoundly
effective in all the scenarios we have evaluated including both white-box and black-box settings.

\section{RELATED WORK}\label{sec:related works}

There are a few recent 
works that focus on generating adversarial examples beyond the small $L_p$-norm perturbation restriction.  In one direction, novel perturbation measures beyond $L_p$ norms such as Wasserstein distance have been proposed~\cite{wong2019wasserstein}. 
In a different direction, 
small perturbations are imposed onto a latent representation of images instead of the images themselves~\cite{zhao2017generating,song2018generative,qiu2019semanticadv}. Further,  a growing line of work 
exploit domain knowledge to 
relax the small perturbation restriction while keeping the generated images natural and meaningful. Our work is aligned with this general framework. 
In particular,
the work in~\cite{an2019big} focuses on exploiting texture transfer and colorization to generate unrestricted images.
In the context of physical-world attacks, perceivable perturbations that resemble real and inconspicuous objects have been proposed~\cite{eykholt2017robust,sharif2016accessorize}.

Our SPA approach extends the Structure-Preserving Transformation (SPT) technique in our previous work~\cite{SPT}, where we
define a structure as a set of all pixels with the \textit{same} pixel value, which is a special case of our definition when each interval contains a single pixel value.
As SPT completely abandons the perturbation restriction, 
the uncontrolled excessive distortion
  may lead to unnatural adversarial examples. This drawback has been staved off in our SPA.
There are a few recent works that also consider structural-aware perturbations similar to ours. 
In particular, color-shifted images are proposed in~\cite{hosseini2018semantic} where 
RGB images are first converted into the HSV color space and the hue and saturation components are then changed randomly where all the pixels in the same channel are perturbed by the same amount. This can be viewed as a special case of our approach where the pixel value partition has a single interval. 
In~\cite{xu2018structured}, 
structural perturbations are generated by penalizing the so called {\it group sparsity}.  
In contrast, our definition of structure patterns is better aligned with human intuition. Moreover, the attack in~\cite{xu2018structured} can be viewed as a specific $L_{\infty}$-bounded attack with an additional strong group sparsity restriction. Thus, it still suffers from the shortcomings of small-perturbation based attacks.

\section{CONCLUSION}

In this paper, we propose structure-preserving attack
(SPA) as a new technique for generating natural and highly transferable adversarial examples. 
SPA is built upon an intuitive definition of structure patterns and introduces the concept of structural perturbation that relaxes the traditional small-perturbation requirement.
Empirical results on the MNIST and CIFAR10 datasets show that SPA 
exhibits strong attack ability in both the white-box and black-box settings even when defenses are applied. Further, when combined with PGD and CW attacks, SPA+PGD and SPA+CW exhibit even stronger white-box attack ability 
while retaining the good transferability of SPA.  
 

We analyze the attack abilities of SPA and baseline attacks in terms of their space flexibility and distortion flexibility. 
The key insight is that it is beneficial to allow more distortion flexibility at the cost of 
space flexibility in order to achieve higher attack ability. We highlight that the high successful attack rates and the outstanding transferability of SPA stem from  
the fact that SPA 
exhibits greater distortion flexibility compared with traditional small-perturbation based approaches. By bridging the gap between the attacks that follow the strict small-perturbation restriction (extremely low distortion flexibility and extremely high space flexibility) and the attacks that allow unbounded distortions (extremely high distortion flexibility and low space flexibility), SPA opens up a new direction on generating natural and strong adversarial examples.

\bibliography{ref}

\end{document}